\documentclass[conference]{IEEEtran}
\IEEEoverridecommandlockouts
\usepackage{cite}
\usepackage{amsmath,amssymb,amsfonts}
\usepackage{algorithm}
\usepackage{algorithmic}
\usepackage{graphicx}
\usepackage{subfig}
\usepackage{textcomp}
\usepackage{xcolor}
\usepackage{booktabs}
\usepackage{url}

\def\BibTeX{{\rm B\kern-.05em{\sc i\kern-.025em b}\kern-.08em
    T\kern-.1667em\lower.7ex\hbox{E}\kern-.125emX}}
\begin{document}

\title{Deep Generative Classifiers for Thoracic Disease Diagnosis with Chest X-ray Images
\thanks{This work was supported in part by NIH Grant 1R21LM012618-01.}
}

\author{\IEEEauthorblockN{Chengsheng Mao}
\IEEEauthorblockA{\textit{Dept. of Preventive Medicine} \\
\textit{Northwestern University}\\
Chicago, IL, USA \\
chengsheng.mao@northwestern.edu}  \\
\IEEEauthorblockN{Liang Yao}
\IEEEauthorblockA{\textit{Dept. of Preventive Medicine} \\
\textit{Northwestern University}\\
Chicago, IL, USA  \\
liang.yao@northwestern.edu}  
\and
\IEEEauthorblockN{Yiheng Pan}
\IEEEauthorblockA{\textit{Dept. of EECS} \\
\textit{Northwestern University}\\
Chicago, IL, USA  \\
yihengpan2019@u.northwestern.edu}  \\
\IEEEauthorblockN{Yuan Luo (Corresponding)}
\IEEEauthorblockA{\textit{Dept. of Preventive Medicine} \\
\textit{Northwestern University}\\
Chicago, IL, USA  \\
yuan.luo@northwestern.edu}  
\and
\IEEEauthorblockN{Zexian Zeng}
\IEEEauthorblockA{\textit{Dept. of Preventive Medicine} \\
\textit{Northwestern University}\\
Chicago, IL, USA  \\
zexian.zeng@northwestern.edu} 

% * <yuan.hypnos.luo@gmail.com> 2018-09-01T15:38:07.521Z:
% 
% Is it possible to adjust the left indentation for second row of authors, e.g., move Liang part a bit to the right
% 
% ^.

% \and
}

\maketitle

\begin{abstract}
Thoracic diseases are very serious health problems that plague a large number of people. Chest X-ray is currently one of the most popular methods to diagnose thoracic diseases, playing an important role in the healthcare workflow. However, reading the chest X-ray images and giving an accurate diagnosis remain challenging tasks for expert radiologists. With the success of deep learning in computer vision, a growing number of deep neural network architectures were applied to chest X-ray image classification. However, most of the previous deep neural network classifiers were based on deterministic architectures which are usually very noise-sensitive and are likely to aggravate the overfitting issue. In this paper, to make a deep architecture more robust to noise and to reduce overfitting, we propose using deep generative classifiers to automatically diagnose thorax diseases from the chest X-ray images. Unlike the traditional deterministic classifier, a deep generative classifier has a distribution middle layer in the deep neural network. A sampling layer then draws a random sample from the distribution layer and input it to the following layer for classification. The classifier is generative because the class label is generated from samples of a related distribution. Through training the model with a certain amount of randomness, the deep generative classifiers are expected to be robust to noise and can reduce overfitting and then achieve good performances. We implemented our deep generative classifiers based on a number of well-known deterministic neural network architectures, and tested our models on the chest X-ray14 dataset. The results demonstrated the superiority of deep generative classifiers compared with the corresponding deep deterministic classifiers.    
% * <yuan.hypnos.luo@gmail.com> 2018-09-01T16:17:21.744Z:
% 
% > The classifier is generative because the class label is generated from samples of a related distribution.
% If you mean P(y|x), isn't it the classical definition of discriminative model?
% 
% ^.
\end{abstract}


\section{Introduction}
Thoracic diseases encompass a variety of serious illnesses and morbidities with high prevalence, e.g. pneumonia affect millions of people worldwide each year and about 50,000 people die from pneumonia each year in the United States only \cite{cdc2017pneumonia}. Detecting the thoracic diseases early and correctly can help clinicians to improve patient treatment effectively. Chest X-ray (CXR), also known as  chest radiograph, is a projection radiograph of the chest and is used to diagnose conditions affecting the chest, its contents, and nearby structures. Due to its affordable price and quick turnaround, CXR is currently one of the most popular radiological examinations to diagnose thoracic diseases. Currently, reading CXR and giving an accurate diagnosis rely on expert knowledge and medical experience of radiologists. With the increasing amount of CXR images, to handle the heavy and tedious workload of reading the CXR images with subtle texture changes, even the most experienced expert may be prone to make mistakes. Therefore, it is important to develop a Computer-Aided Diagnosis (CAD) system to automatically detect different types of thoracic diseases by reading patients' CXR images.

Developing a CAD system to understand medical images and to diagnose diseases has attracted wide research interests for decades \cite{murphy2009large,chen2011development,papadopoulos2005characterization}. However, traditional statistical learning methods, such as Bayesian classifier \cite{domingos1996beyond,hu2015bayesian,wang2014non}, SVM \cite{cortes1995support,fan2008liblinear,chang2011libsvm} and KNN \cite{larose2006k,cover1967nearest,mao2015nearest,mao2015learning} etc., are not expert in directly handling the medical images in the high-dimensional pixel-level features. They usually require onerous highly customized feature engineering before classification, thus, they cannot generalize well and is expert labor intensive. With the success of deep learning in computer vision, it is natural to apply deep learning models to assist in disease diagnosis based on medical images. Recently, deep learning based CAD has benefited many biomedical applications \cite{litjens2017survey}, e.g., diabetic eye disease diagnosis \cite{gulshan2016development}, cancer metastases detection and localization \cite{liu2017detecting}, lung nodule detection \cite{setio2017validation}, survival analysis \cite{zhu2017wsisa} and clinical notes classification \cite{luo2017segment,luo2017recurrent}, etc. In this work, we develop a generative deep neural network architecture and apply it to diagnosing thoracic diseases based on CXR images.
% * <yuan.hypnos.luo@gmail.com> 2018-09-01T16:36:50.537Z:
% 
% > are rarely feasible to directly handle the medical images in the high-dimensional pixel-level features.
% I don't think it's fair to say so, CV and image processing has long history of feature engineering (e.g. sift features), mainly to improve the performance rather than to make the computation feasible. You can say from the perspective that too many highly customized feature engineering do not generalize well and is expert labor intensive, hence a drawback. 
% 
% ^.

Deep neural networks usually require large-scale datasets for training. Recently, Wang et al. \cite{wang2017chestxray} released the datasets ChestX-ray8 and later ChestX-ray14 which is considered one of the largest public chest X-ray dataset (details in Section \ref{sec:dataset}). There are 14 thoracic diseases included in ChestX-ray14, i.e., Atelectasis, Cardiomegaly, Effusion, Infiltration, Mass, Nodule, Pneumonia, Pneumothorax, Consolidation, Edema, Emphysema, Fibrosis, Pleural Thickening, and Hernia. We focus on this  dataset to train a deep generative classifier to diagnose the 14 diseases.
% * <yuan.hypnos.luo@gmail.com> 2018-09-01T16:48:52.518Z:
% 
% > Deep neural networks
% Because deep learning is such a buzzword, we should minimize its usage in the paper, deep neural networks sounds better, and wherever possible, use more specific names, such as CNN, RNN, LSTM etc.
% 
% ^.

\begin{figure}[tbp]
\includegraphics[width=0.5\textwidth,page=1]{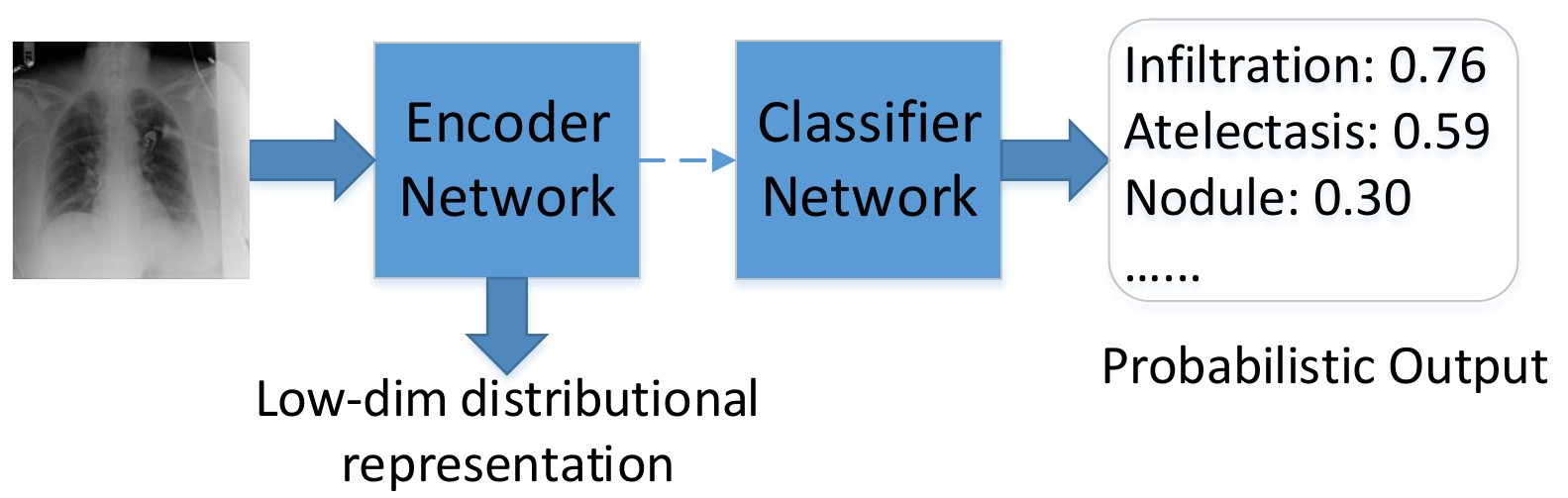}
\caption{An overview of the deep generative classifier for thorax disease diagnosis. The network reads chest X-ray images and produces a probability score for each thorax disease. The dashed arrow between the encoder network and the classifier network denotes the two networks are connected by random sampling.}
\label{fig:architecture}
\end{figure}

In this paper, we propose using deep generative classifiers to automatically diagnose thorax diseases from CXR images. A Deep Generative Classifier (DGC) contains an encoder network and a classifier network. The encoder network encodes each input CXR image to a low-dimensional distribution of latent features. The classifier network classifies a sample using features drawn from the latent low-dimensional distribution and outputs probabilities of class label assignment. Our main idea is to use the random sampling connection between the encoder network and the classifier network rather than a deterministic connection which is adopted in most previous frameworks, e.g. AlexNet \cite{krizhevsky2012imagenet}, ResNet \cite{he2016deep}, VGGNet \cite{simonyan2014very} and DenseNet \cite{huang2017densely}. Intuitively, by using the random sampling connection, the model would be more robust to noise and can reduce overfitting. An overview of our generative classifier is shown in Fig. \ref{fig:architecture}. As shown in Fig. \ref{fig:architecture}, given a CXR image, our model will output a list of probabilities corresponding to a list of thorax diseases, and a low-dimensional distributional representation of this image as a by-product which can be used for other classification or clustering tasks. 
% * <yuan.hypnos.luo@gmail.com> 2018-09-01T16:56:36.214Z:
% 
% > The encoder network encodes each input CXR image to a low-dimensional distribution of latent features. The classifier network classifies a sample using features drawn from the latent low-dimensional distribution and outputs probabilities of class label assignment.
% I rephrased it in terms of drawing features from distribution, let me know if this is not what you did.
% 
% ^.

\section{Related Work}

\subsection{Deep Learning for CXR Image Analysis}
Many research efforts have been directed towards automatic detection of thorax diseases based on diverse data generated by chest X-ray scanning. Before ChestX-ray14 dataset was released, there were also some works about thoracic disease classification on some relatively small dataset. Bar et al. \cite{bar2015chest} applied a pre-trained Decaf Convolutional Neural Network (CNN) model \cite{donahue2014decaf} to classify 8 thoracic diseases on a dataset of 433 images. Lajhani et al. \cite{lakhani2017deep} ensembled both AlexNet \cite{krizhevsky2012imagenet} and GoogleNet \cite{szegedy2015going} for tuberculosis classification on  a dataset of 1007 posteroanterior
CXR images.
% * <yuan.hypnos.luo@gmail.com> 2018-09-01T17:11:25.716Z:
% 
% >  Bar et al. \cite{bar2015chest} applied a pre-trained Decaf Convolutional Neural Network (CNN) model \cite{donahue2014decaf} to classify 8 thoracic diseases. Lajhani et al. \cite{lakhani2017deep} ensembled both AlexNet \cite{krizhevsky2012imagenet} and GoogleNet \cite{szegedy2015going} for tuberculosis classification. The two research showed promising results on relatively small dataset (less than 500 images).
% What's your criteria for selecting these two from the many? 
% 
% ^.

Since Wang et al. \cite{wang2017chestxray} released the datasets ChestX-ray14, there has been an increasing amount of research on CXR analysis using deep neural networks. Wang et al. \cite{wang2017chestxray} also evaluated the performance of four classic deep learning architectures (i.e., AlexNet \cite{krizhevsky2012imagenet}, VGGNet \cite{simonyan2014very}, GoogLeNet \cite{szegedy2015going} and ResNet \cite{he2016deep}) to diagnose 14 thoracic diseases from CXR images. To explore the correlation among the 14 diseases, Yao et al. \cite{yao2017learning} used a Long-short Term Memory (LSTM) \cite{hochreiter1997long}  to repeatedly decode the feature vector from a DenseNet \cite{huang2017densely} and produced one disease prediction at each step. Kumar et al. \cite{kumar2018boosted} explored suitable loss functions to train a convolutional neural network (CNN) from scratch and presented a boosted cascaded CNN for multi-label CXR classification. Rajpurkar et al. \cite{Rajpurkar2017CheXNet} achieved good multi-label classification results by fine-tuning a pre-trained DenseNet121 \cite{huang2017densely}. Li et al. \cite{li2017thoracic} used a pre-trained ResNet to extract features and divided them into patches which are passed through a fully convolutional network (FCN) \cite{long2015fully} to obtain a disease probability map. 
% * <yuan.hypnos.luo@gmail.com> 2018-09-01T17:17:17.642Z:
% 
% > Further related research includes \cite{guan2018diagnose,zhou2018weakly,yan2018weakly}. 
% What's the criteria for putting those together? Weakly supervised models?
% 
% ^.
% * <yuan.hypnos.luo@gmail.com> 2018-09-01T17:15:53.007Z:
% 
% > a fully-convolutional classification CNN
% What is this?
% 
% ^.

Previous deep learning architectures all had a deterministic mapping between encoded features and CXR classification. In this paper, we propose using deep generative classifiers for classifying thoracic disease with CXR images. By introducing a generative process, the learned DGC model should be more robust to noise and reduce the overfitting issue.
% * <yuan.hypnos.luo@gmail.com> 2018-09-01T17:19:17.694Z:
% 
% > In this paper, we propose using deep generative classifiers for classifying thoracic disease using CXR images.
% I rephrased the sentence, in general, we want to avoid the claims that we build a model to diagnose a disease (e.g., replace physicians).
% 
% ^.

\subsection{Variational Autoencoder}
The deep generative classifiers have similar traits with Variational Auto-Encoder (VAE) \cite{Kingma2013auto}. In VAE, a high-dimensional sample is encoded to a low-dimensional feature distribution, a sample from this distribution is decoded to the original high-dimensional sample. To generate new samples better, VAE needs to constrain the low-dimensional distribution to a certain known distribution. VAE was usually used to reduce the data dimension or to generate new samples, but was rarely used for supervised classification directly. In this paper, we leverage VAE to design a generative classification model where the class label was generated from the latent low-dimensional distribution.
% * <yuan.hypnos.luo@gmail.com> 2018-09-01T17:21:23.092Z:
% 
% > have similar traits
% avoid using "something" too informal.
% 
% ^.

\section{Models}
Our purpose is to judge whether one or more thoracic diseases are presented in a CXR image, it can be modeled as a multi-label classification problem. We integrate the losses of the multiple objectives for multiple labels, and tackle this problem using deep generative classifiers.
% * <yuan.hypnos.luo@gmail.com> 2018-09-01T19:08:30.492Z:
% 
% >  it can be modeled as a multi-label classification problem.
% You modeled it as multiple binary classification problem
% 
% ^ <yuan.hypnos.luo@gmail.com> 2018-09-01T19:43:42.118Z.
In this section, we explicitly describe the technical details of the proposed deep generative classifiers. First, we present the detailed architecture of the deep generative classifier. Then, we explain the training strategy of the model. Finally, we give a probabilistic interpretation of our model.

\begin{figure*}[tbp]
\includegraphics[width=\textwidth,page=2]{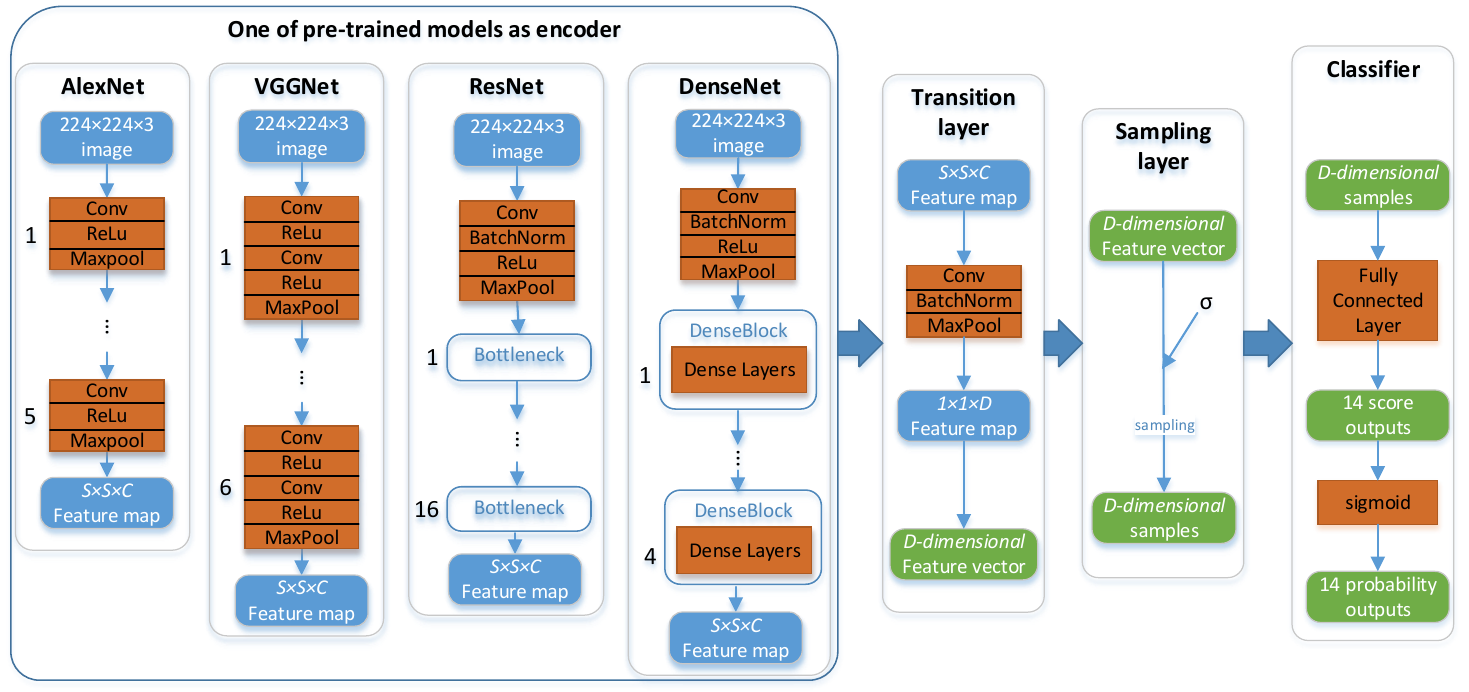}
\caption{The framework of deep generative classifiers.}
\label{fig:detail}
\end{figure*}

\subsection{Architecture}
A DGC receives a batch of CXR image as input and computes a list of probabilities for each disease. In the framework, each input image is encoded to a latent low-dimensional distribution by the encoder network, the classifier network classifies the sample based on features drawn from the latent low-dimensional distribution to generate a probabilistic output. Fig. \ref{fig:detail} illustrates the detailed framework of DGC. Given a CXR image input $X$, the computation flows through a series of sub-modules, including the encoder, the transition layer, the sampling layer, and the classifier. Next, we explain the sub-modules in detail.

\subsubsection{Encoders}  \label{sec:encoder}
As shown in Fig. \ref{fig:detail}, the encoder in our network is leveraged from a part of a pre-trained model on ImageNet \cite{russakovsky2015imagenet,deng2009imagenet}, e.g. AlexNet \cite{krizhevsky2012imagenet}, VGGNet \cite{simonyan2014very}, ResNet \cite{he2016deep} and DenseNet \cite{huang2017densely}. For the pre-trained models, we discard the fully-connected layers and classification layers, and keep the feature layers to extract feature maps for CXR images. Through the encoder, an original image $X$ ($224 \times 224 \times 3$) is encoded to $C$ feature maps with size $S \times S$, represented by $En(X;\Phi_e)$ where $\Phi_e$ is the set of trainable parameters of the encoder.  

\subsubsection{Transition Layer}
The transition layer is to transform the output feature maps of the encoder to a flat feature vector that has a uniform dimension for different images. Due to the variety of the output feature maps of the encoder we adopt, we use a convolution layer with a certain number ($D$) of filters to get $D$ feature maps, and then adopt batch normalization \cite{ioffe2015batch} right after the convolution layer to ease the training process. Finally, a max pooling layer \cite{boureau2010theoretical} with kernel size equal to the feature map size is applied to reduce each feature map to $1 \times 1 \times D$. The $1 \times 1 \times D$ feature map is then squeezed to a $D$-dimensional feature vector, represented by 
\begin{equation}
\mu=Trans(En(X;\Phi_e);\Phi_t)
\label{eq:transition}
\end{equation}
where $\Phi_t$ is the set of trainable parameters of the transition layer.

\subsubsection{Sampling Layer}
Instead of using the output of the transition layer directly as features, we treat it as the mean of a latent distribution. We let this sampling layer samples from this latent distribution to produce a feature vector. In our design, the features of a CXR image adopt a latent distribution whose parameters are computed from the upstream networks or jointly learned with the network parameters by Stochastic Gradient Descent (SGD) \cite{sutskever2013importance}. In our architecture, we treat the latent distribution as Gaussian distributions where the mean is computed from the upstream network, i.e., the output of the transition layer, and the covariance matrix is diagonal with the diagonal elements learned with the network parameters. Thus, for a $D$-dimensional latent Gaussian distribution, the sampling layer has a total of $D$ parameters each of which corresponds to the standard deviation of each dimension. The output of sampling layer is a feature vector sampled from the latent distribution. The sampling layer is crucial to make the model generative rather than deterministic. The sampling layer can be represented by 
% * <yuan.hypnos.luo@gmail.com> 2018-09-01T19:41:15.277Z:
% 
% > generative
% changed probabilistic to generative, as even the Alex net can be regarded as probabilistic as the softmax layer gives out probabilities. 
% 
% ^.
% * <yuan.hypnos.luo@gmail.com> 2018-09-01T19:39:44.467Z:
% 
% > The output of sampling layer is a feature vector sampled from the latent distribution.
% It's problematic to say you redraw the sample, this means you are classifying a different sample. Instead, I rephrased it as redraw the feature vector of that sample.
% 
% ^.
% * <yuan.hypnos.luo@gmail.com> 2018-09-01T19:37:52.360Z:
% 
% > with the network parameters.
% again, which networks, encoder or classification?
% 
% ^.
% * <yuan.hypnos.luo@gmail.com> 2018-09-01T19:36:16.500Z:
% 
% > jointly learned with the network
% which networks, encoder or classification?
% 
% ^.
\begin{equation}
z \thicksim N(\mu,\sigma)
\label{eq:sampling}
\end{equation}
where $\mu$ is the output of the transition layer and $\sigma$ is the trainable parameter of the sampling layer.

\subsubsection{Classifier}
Since we use a low-dimensional representation as the feature vector of an input CXR image, we need a classifier to discriminate if a disease is presented. For the 14 thoracic diseases, there are 14 probabilist outputs considered as a 14-dimensional vector. In our architecture, we use a fully-connected layer with 14 score outputs that are then transformed by a sigmoid function to probabilistic outputs. The classifier is represented by 
\begin{equation}
p=f(z;\Phi_c)
\label{eq:classifier}
\end{equation}
where $z$ is the output of the sampling layer and $\Phi_c$ is the parameters of the classifier.

\subsection{Training Strategy}
We next describe two crucial strategies related to training that concern the loss function for multi-label classification and the reparameterization for distribution parameter learning.

\subsubsection{Loss Function}
To train our network, we must define a loss function for the multiple outputs corresponding to the 14 diseases. The true label of each image is considered a 14-dimensional vector $y=[y_1,\cdots,y_i,\cdots, y_C], y_i \in \{0,1\}, C=14$ where $y_i$ represents whether the corresponding disease is presented, i.e., 1 for presence and 0 for absence. An all-zero vector
represents ``No Findings'' in the 14 diseases. We compute the cross entropy loss $l_i$ for disease $i$ as \eqref{eq:crossentropy}, where $p_i$ is the output probability of disease $i$.  
\begin{equation}
\begin{aligned}
l_i &= -y_i*\ln p_i -(1-y_i)*\ln (1-p_i)   \\
    &=
    \begin{cases}
        -\ln p_i, \quad ~\quad \qquad  y_i=1;  \\
        -\ln (1-p_i),  \qquad   y_i=0
    \end{cases}
\end{aligned}
\label{eq:crossentropy}
\end{equation}

For a mini-batch with $n$ samples, the corresponding targets are $n$ $C$-dimensional (0,1)-vectors which can be considered as a (0,1)-matrix with shape $n \times C$ which we call target matrix. Since there usually are only a few pathologies present in a CRX image, the target matrix should be a sparse matrix where there are many more `0's than `1's. To balance the influence of `0's and `1's on the loss, we weight the losses for different classes. In the mini-batch, we design the weights as \eqref{eq:weight}, where $|P|$ and $|N|$ are, respectively, the number of `1's and `0's in the target matrix of the mini-batch. Thus, combining \eqref{eq:crossentropy} and \eqref{eq:weight}, we define the loss function for a CRX image as \eqref{eq:loss}
% * <yuan.hypnos.luo@gmail.com> 2018-09-01T20:48:54.852Z:
% 
% > pathologies
% I noticed you sometimes use disease, sometimes use pathology, need to be consistent.
% 
% ^.
\begin{equation}
w_i =
    \begin{cases}
        \frac{|P|+|N|}{|P|},  \qquad  y_i=1;  \\
        \frac{|P|+|N|}{|N|},  \qquad  y_i=0
    \end{cases}
\label{eq:weight}
\end{equation}

\begin{equation}
\begin{aligned}
L(p,y) &= \sum_{i=1}^{14}{w_i l_i}     \\
       &=-\frac{|P|+|N|}{|P|}\sum_{y_i=1}{\ln {p_i}}-\frac{|P|+|N|}{|N|}\sum_{y_i=0}{\ln (1-p_i)}
\end{aligned}
\label{eq:loss}
\end{equation}

\subsubsection{Reparameterization Trick}
Since the sampling operation is not differentiable, we use the reparameterization trick \cite{Kingma2013auto}. In our architecture, the latent distribution is Gaussian (assumed $N(\mu,\sigma^2)$). Due to random sampling, the derivative of a sample $z$ from this distribution with respect to $\mu$ and $\sigma$ cannot be directly obtained. To learn the parameters with SGD, we construct a deterministic relation between $z$ and $\mu,\sigma$ by introducing an auxiliary variable $\epsilon$, $\epsilon \thicksim N(0,1)$. Thus, sampling from $N(\mu,\sigma^2)$ is equal to sampling $\epsilon$ from $N(0,1)$ and then computing the sample by 
\begin{equation}
z=\mu+\epsilon \sigma, \quad  \epsilon \thicksim N(0,1)
\label{eq:reparameterization}
\end{equation}

The derivative of $z$ can then be easily computed as 
\begin{equation}
\begin{aligned}
\partial z/\partial \mu =1  \\
\partial z/\partial \sigma =\epsilon
\end{aligned}
\label{eq:gradient}
\end{equation}

Combining \eqref{eq:transition}, \eqref{eq:sampling}, \eqref{eq:reparameterization}, \eqref{eq:classifier}, we can compute the output of the model by \eqref{eq:dgc}. Thus, through the reparameterization trick, the network can be trained by SGD. 
\begin{equation}
p= f(Trans(En(X;\Phi_e);\Phi_t)+\epsilon \sigma;\Phi_c)
\label{eq:dgc}
\end{equation}

\subsubsection{Training Algorithm}
The training algorithm of DGC is briefly described in Algorithm \ref{al:train}. It starts by initializing the model parameter $\Phi$ (Line \ref{al:init}), then repeats the training procedure until the terminating condition (e.g., a threshold number of epochs or early stop) is met (Lines \ref{al:repeat}-\ref{al:until}). In an epoch in the training procedure, the training set is split into batches (Line \ref{al:split}). For each batch in the training set (Line \ref{al:batch}-\ref{al:endbatch}), we compute the loss of the batch (Lines \ref{al:count}-\ref{al:endsample}), and the gradients with respect to the parameters $\Phi$ for error backpropagation (Line \ref{al:backprpgation}). We then update the parameters $\Phi$ using the gradients and a certain optimization algorithm (Line \ref{al:update}). Finally, if the terminating condition is met, the model is updated with parameters $\Phi$. 
% * <yuan.hypnos.luo@gmail.com> 2018-09-01T21:00:40.255Z:
% 
% > terminating condition
% We usually specify what the terminating condition is, also in the algorithm
% 
% ^ <yuan.hypnos.luo@gmail.com> 2018-09-01T21:21:18.452Z.

\begin{algorithm}[t]  
\caption{ Deep Generative Classifier Training. $N$ is the number of samples in the training set. The parameter list $\Phi=[\Phi_e,\Phi_t,\sigma,\Phi_c]$. }  
\label{al:train}
\begin{algorithmic}[1] 
\REQUIRE ~~ \\
Training set $Tr=\{(X_i,Y_i)\}, i\in \{1,\cdots,N\}$ ; \\
Batch size $|B|$;    \\
The terminating condition (e.g., a number of epochs or early stop).   \\
\ENSURE The updated parameter list of DGC $\Phi$.
\STATE Initialize the parameter list of DGC $\Phi$      \label{al:init}
\REPEAT  \label{al:repeat}
\STATE Randomly split the training set $Tr$ into $n$ mini-batches $\{B_1,\cdots,B_n\}$ with each batch has $|B|$ samples, the last batch can have less samples   \label{al:split}

\FOR { $B$ in $\{B_1,\cdots,B_n\}$ }  \label{al:batch}
\STATE Count the number of `0's $|N|$ and number of `1's $|P|$ in the target matrix $Y$  \label{al:count}
\STATE $\mathcal{L}=0$  \label{al:initloss}
\FOR {$(X,y)$ in $B$}  \label{al:sample}
\STATE Random sampling $\epsilon$ with the same shape as $\sigma$ from $N(0,1)$  \label{al:sampling}
\STATE Compute the output $p$ of the model by \eqref{eq:dgc}  \label{al:output}
\STATE Compute the loss $L$ by \eqref{eq:loss}   \label{al:loss}
\STATE $\mathcal{L}=\mathcal{L}+L$  \label{al:batchloss}
\ENDFOR  \label{al:endsample}
\STATE Compute the gradient $g=\nabla_\Phi \mathcal{L}$  \label{al:backprpgation} 
\STATE Update parameters $\Phi$ using gradients $g$, $\Phi=update(\Phi,g)$. e.g. SGD \cite{sutskever2013importance} or Adam \cite{kingma2014adam}  \label{al:update}
\ENDFOR  \label{al:endbatch}
\UNTIL the terminating condition is met  \label{al:until}

\RETURN $\Phi$     \label{al:return}

\end{algorithmic}  
\end{algorithm} 

\subsection{Test Strategy}
After the model is trained, in the test process, a CXR image can be encoded to a Gaussian distribution $N(\mu, \sigma)$. To classify the image, we need sampling the feature vector of a sample from the distribution, and then input the sample to the classifier to get the probability output. However, the sampling process is random, that is, the output may change with the different samplings. Thus, to achieve a consistent classification output in the test process, we use the expectation of the distribution as the sample input to the classifier, i.e.,
% * <yuan.hypnos.luo@gmail.com> 2018-09-01T21:03:15.127Z:
% 
% > To achieve a stable classification output, we use the expectation of the distribution as the sample input to the classifier, 
% I am confused by this description. Doesn't this defeat the purpose of having a distribution? Also, this \mu is probably different from the feature maps from encoder that you used as mean for low-dimensional feature distribution, right? If so, need to use different symbols, or even \mu'
% 
% ^.
\begin{equation}
z=\mathbb{E}N(\mu,\sigma)=\mu
\label{eq:expectation}
\end{equation}

\section{Experiments}
In this section, we evaluate the performance of the proposed DGC model and compared it with the corresponding deep deterministic classifiers. 
\subsection{ChestX-ray14 Dataset} \label{sec:dataset}
We evaluate and validate the DGC model using the ChestX-ray14 dataset\footnote{https://nihcc.app.box.com/v/ChestXray-NIHCC\label{fn:dataurl}} \cite{wang2017chestxray}. The ChestX-ray14 dataset consists of 112,120 frontal-view chest X-ray images of 30,805 unique patients. There are 14 thoracic disease labels included in these images (i.e., Atelectasis, Cardiomegaly, Effusion, Infiltration, Mass, Nodule, Pneumonia, Pneumothorax, Consolidation, Edema, Emphysema, Fibrosis, Pleural Thickening and Hernia). The labeled ground truth is obtained through Natural Language Processing (NLP) on the patients' diagnostic reports, the labeling accuracy is estimated to be $>90\%$ \cite{wang2017chestxray}. Out of the 112,120 CXR images, 51,708 contains one or more pathologies. The remaining 60,412 images are considered normal. In our experiments, we split the dataset into training-validation set and test set on the patient level using the publicly available data split list\textsuperscript{\ref{fn:dataurl}}. All studies from the same patient will only appear in either training-validation set or testing set. The detailed information about the number of images and patients in training-validation set and test set is shown in Table \ref{tab:datasetsummary}.

\begin{table}[tbp]
%   \centering
  \caption{Number of images and patients in the ChestX-ray14 dataset}
    \begin{tabular}{l|cc|cc|cc}
    \toprule
          & \multicolumn{2}{c|}{\textbf{Entire Set}} & \multicolumn{2}{c|}{\textbf{Train-val Set}$^2$} & \multicolumn{2}{c}{\textbf{Test Set}} \\
          \cline{2-7}
    \textbf{Diseases} & \multicolumn{1}{l}{\textbf{\textit{\#imgs}}$^3$} & \multicolumn{1}{l|}{\textbf{\textit{\#pts}}$^4$} & \multicolumn{1}{l}{\textbf{\textit{\#imgs}}} & \multicolumn{1}{l|}{\textbf{\textit{\#pts}}} & \multicolumn{1}{l}{\textbf{\textit{\#imgs}}} & \multicolumn{1}{l}{\textbf{\textit{\#pts}}} \\
	\midrule    
    Atelectasis & 11535 & 4974  & 8280  & 4182  & 3255  & 792 \\
    Cardiomegaly & 2772  & 1565  & 1707  & 1228  & 1065  & 337 \\
    Consolidation & 4667  & 2150  & 2852  & 1617  & 1815  & 533 \\
    Edema & 2303  & 1073  & 1378  & 747   & 925   & 326 \\
    Effusion & 13307 & 4273  & 8659  & 3502  & 4648  & 771 \\
    Emphysema & 2516  & 1046  & 1423  & 762   & 1093  & 284 \\
    Fibrosis & 1686  & 1260  & 1251  & 1003  & 435   & 257 \\
    Hernia & 227   & 134   & 141   & 102   & 86    & 32 \\
    Infiltration & 19870 & 8031  & 13782 & 7111  & 6088  & 920 \\
    Mass  & 5746  & 2550  & 4034  & 2115  & 1712  & 435 \\
    Nodule & 6323  & 3390  & 4708  & 2855  & 1615  & 535 \\
    PT$^1$ & 3385  & 2006  & 2242  & 1559  & 1143  & 447 \\
    Pneumonia & 1353  & 955   & 876   & 697   & 477   & 258 \\
    Pneumothorax & 5298  & 1484  & 2637  & 1080  & 2661  & 404 \\
    Normal & 60412 & 16405 & 50500 & 14892 & 9912  & 1513 \\
    \midrule
    Summary & 112120 & 30805 & 86524 & 28008 &	25596 &	2797  \\

    \bottomrule
    \end{tabular}  \newline \\%
    $^1$PT --- Pleural Thickening; $^2$Train-val set --- Training-validation set; $^3$\#imgs --- the number of CXR images; $^4$\#pts --- the number of patients. 
  \label{tab:datasetsummary}%
\end{table}%

The training-validation set is further randomly split into a training set and a validation set, 7/8 as the training set and 1/8 as the validation set. The training set is used to train the model and the validation set is used to select a model according to the performance.  

\subsection{Preprocessing}
Since the ImageNet pre-trained models only accept 3-channel images with size $224 \times 224$, while the images in  ChestX-ray14 dataset are $1024 \times 1024$ grayscale, we convert the grayscale images to 3-channel RGB images, downscale the original resolution to $256\times 256$ and then crop the image to $224 \times 224$ at the center. We normalized the images by mean ($[0.485, 0.456, 0.406]$) and standard deviation ($[0.229, 0.224, 0.225]$) according to the images from ImageNet. We do not apply any data augmentation techniques.
% * <yuan.hypnos.luo@gmail.com> 2018-09-01T21:10:39.540Z:
% 
% > we expand the grayscale image to 3-channel
% Does this mean you duplicate one channel to 3? If so, probably don't want to mention it (it's not wrong, but definitely inefficient) ...
% 
% ^.

\subsection{Experimental setting}
Our experimental setting includes the following aspects.
\subsubsection{Encoder} 
In our experiments, we tried 6 pretrained models as the encoder in our architecture, including AlexNet \cite{krizhevsky2012imagenet}, VGGNet16 \cite{simonyan2014very}, ResNet50 \cite{he2016deep} and DenseNet121 \cite{huang2017densely}, DenseNet161 \cite{huang2017densely}, DenseNet201 \cite{huang2017densely}. As described in Section \ref{sec:encoder}, we discarded the high-level fully-connected layers and classification layers of the pretrained models and only used the feature layers as the encoder. As shown in Fig. \ref{fig:detail}, different encoders have different inner structure and have different parameters. We respectively denote the DGC based on these encoder as DGC-AlexNet (G-AN), DGC-VGGNet16 (G-VN16), DGC-ResNet50 (G-RN50), DGC-DenseNet121 (G-DN121), DGC-DenseNet161 (G-DN161) and DGC-DenseNet201 (G-DN201).
\subsubsection{Baselines} \label{sec:baselines}
The specific part of our architecture is the sampling layer which improves the model robustness to noise. Thus, we remove the sampling layer in the architecture and consider the remaining parts as the baselines, i.e., the output of the transition layer is directly input to the classifier in baselines. Correspondingly, we have 6 baselines, respectively denoted as AlexNet (AN), VGGNet16 (VN16), ResNet50 (RN50), DenseNet121 (DN121), DenseNet161 (DN161) and DenseNet201 (DN201).

\begin{table*}[tbp]
%   \centering
  \caption{comparison of AUC scores between DGCs and the corresponding baselines on ChestX-ray14 dataset. The better results based on the same encoder are \textbf{bolded}. The best results among all the models are colored \textcolor{red}{red}.}
    \begin{tabular}{l|cc|cc|cc|cc|cc|cc}
    \toprule
    Diseases & AN & G-AN & RN50 & G-RN50 & DN201 & G-DN201 & DN121 & G-DN121 & DN161 & G-DN161 & VN16 & G-VN16 \\
    \midrule
    Atelectasis & 0.7327 & \textbf{0.7359} & 0.7407 & \textbf{0.7418} & \textbf{0.7426} & 0.7415 & \textbf{0.7391} & 0.7378 & \textbf{0.7433} & 0.7421 & 0.7488 & \textcolor{red}{\textbf{0.7495}} \\
    Cardiomegaly & 0.8681 & \textbf{0.8699} & 0.8553 & \textbf{0.8623} & 0.8541 & \textbf{0.8606} & 0.8483 & \textbf{0.8517} & \textbf{0.8662} & 0.8653 & \textcolor{red}{\textbf{0.8700}} & 0.8687 \\
    Effusion & 0.7874 & \textbf{0.7896} & 0.7984 & \textbf{0.7991} & 0.7987 & \textbf{0.7997} & \textbf{0.7982} & 0.7976 & 0.8026 & \textbf{0.8052} & 0.8075 & \textcolor{red}{\textbf{0.8096}} \\
    Infiltration & 0.6725 & \textbf{0.6771} & \textbf{0.6798} & 0.6773 & 0.6756 & \textbf{0.6790} & 0.6775 & \textbf{0.6792} & 0.6751 & \textbf{0.6773} & \textcolor{red}{\textbf{0.6894}} & 0.6869 \\
    Mass  & 0.7340 & \textbf{0.7374} & 0.7480 & \textbf{0.7502} & 0.7567 & \textbf{0.7614} & 0.7495 & \textbf{0.7538} & 0.7628 & \textbf{0.7652} & 0.7756 & \textcolor{red}{\textbf{0.7820}} \\
    Nodule & 0.6778 & \textbf{0.6795} & \textbf{0.6962} & 0.6934 & 0.7107 & \textbf{0.7118} & \textbf{0.7032} & 0.7029 & 0.7142 & \textbf{0.7188} & 0.7248 & \textcolor{red}{\textbf{0.7255}} \\
    Pneumonia & 0.6632 & \textbf{0.6744} & \textbf{0.6877} & 0.6868 & 0.6861 & \textbf{0.6917} & \textbf{0.6878} & 0.6858 & 0.6939 & \textcolor{red}{\textbf{0.7020}} & 0.6943 & \textbf{0.6954} \\
    Pneumothorax & 0.8149 & \textbf{0.8195} & 0.8364 & \textbf{0.8405} & 0.8371 & \textbf{0.8396} & 0.8356 & \textbf{0.8405} & 0.8437 & \textcolor{red}{\textbf{0.8497}} & 0.8441 & \textbf{0.8451} \\
    Consolidation & 0.7088 & \textbf{0.7105} & 0.7112 & \textbf{0.7136} & 0.7129 & \textbf{0.7177} & 0.7124 & \textbf{0.7167} & 0.7174 & \textbf{0.7190} & 0.7259 & \textcolor{red}{\textbf{0.7283}} \\
    Edema & 0.8280 & \textbf{0.8311} & 0.8273 & \textbf{0.8311} & 0.8294 & \textbf{0.8302} & \textbf{0.8287} & 0.8280 & 0.8355 & \textcolor{red}{\textbf{0.8368}} & \textbf{0.8363} & 0.8340 \\
    Emphysema & 0.7991 & \textbf{0.8059} & 0.8659 & \textbf{0.8724} & \textbf{0.8704} & 0.8668 & 0.8579 & \textbf{0.8590} & 0.8793 & \textcolor{red}{\textbf{0.8823}} & \textbf{0.8707} & 0.8699 \\
    Fibrosis & 0.7956 & \textbf{0.7977} & \textbf{0.7939} & 0.7885 & 0.7982 & \textbf{0.8036} & 0.7873 & \textbf{0.7905} & 0.7972 & \textcolor{red}{\textbf{0.8103}} & \textbf{0.7979} & 0.7978 \\
    PT$^*$    & 0.7356 & \textbf{0.7373} & 0.7379 & \textbf{0.7424} & 0.7424 & \textbf{0.7432} & \textbf{0.7465} & 0.7457 & 0.7451 & \textbf{0.7510} & 0.7554 & \textcolor{red}{\textbf{0.7581}} \\
    Hernia & 0.8485 & \textbf{0.8498} & \textbf{0.8879} & 0.8816 & 0.8974 & \textcolor{red}{\textbf{0.9113}} & \textbf{0.8951} & 0.8898 & \textbf{0.9097} & 0.9015 & \textbf{0.8837} & 0.8776 \\
    \midrule
    Average  & 0.7619 & \textbf{0.7654} & 0.7762 & \textbf{0.7772} & 0.7794 & \textbf{0.7827} & 0.7762 & \textbf{0.7771} & 0.7847 & \textbf{0.7876} & 0.7875 & \textbf{0.7877} \\
    \bottomrule
    \end{tabular} \newline \\%  
$^*$PT --- Pleural Thickening; AN --- AlexNet; VN16 --- VGGNet16; RN50 --- ResNet50; DN121 --- DenseNet121; DN161 --- DenseNet161; DN201 --- DenseNet201; G-AN --- DGC-AlexNet; G-VN16 --- DGC-VGGNet16; G-RN50 --- DGC-ResNet50; G-DN121 --- DGC-DenseNet121; G-DN161 --- DGC-DenseNet161; G-DN201 --- DGC-DenseNet201.
  \label{tab:results}%
% * <yuan.hypnos.luo@gmail.com> 2018-09-01T21:27:54.002Z:
% 
% > $^*$PT --- Pleural Thickening.
% You also want to explain abbrev. such as AN, conventions such as G-AN. 
% 
% ^.
\end{table*}%

\subsubsection{Hyperparameters} 
% * <yuan.hypnos.luo@gmail.com> 2018-09-01T21:15:13.882Z:
% 
% > Super-parameters
% We usually call them hyperparameters, unless you really meant for something else.
% 
% ^.
\begin{itemize}
\item Initialization: the encoder was initialized with a pretrained model. In the transition layer, the convolution layer was initialized with kaiming uniform initialization \cite{he2015delving}, the batch normalization layer was initialized with weights drawn from the uniform distribution $U(0,1)$ and bias as zero. To ensure the positivity of $\sigma$ in the sampling layer, we computed $\sigma$ from $\sigma^2=e^v$ where $v$ was initialized with values drawn from the uniform distribution $U(0,1)$. In the classifier, the parameters of the fully-connected layer were initialized with kaiming uniform initialization \cite{he2015delving}. 
\item Latent dimension: we set the latent dimension to 1024, i.e., each image was encoded to 1024-dimensional feature vector adopting Gaussian distribution.
\item Batch size: the batch size was 16, i.e., the model updated parameters per 16 images.
\item Optimizer: we trained the model by Adam optimizer \cite{kingma2014adam} with parameters $lr=10^{-5},\beta=(0.9, 0.999), eps=10^{-8}, weight\_decay=0$
\item Terminating condition: we terminated the training procedure when it repeated 10 epochs. In each epoch, we tested the model on the validation set and save the model with the best performance.
\end{itemize}

\subsubsection{Evaluation}
Since our model outputs the probability for each disease, it is natural to plot a Receiver Operating Characteristic curve (ROC) for each disease. In our experiments, we calculated the Area Under ROC (AUC) for each disease, and evaluated the classification performance by the average AUC of the 14 diseases.  

\subsection{Implementation}
Our models and all the baselines were implemented using Python 3.6 with PyTorch framework on a CentOS Linux server. The models were trained and tested on 4 Tesla K40 GPUs. We made our source code publicly available at \url{https://github.com/mocherson/deep-generative-classifiers}

\subsection{Classification Results and Analysis}
In each of the 10 epochs of training, we evaluated the model on the validation set, and selected the model that achieved the highest classification performance to test on the test set. We repeated the training and classification procedure 5 times and reported the average results. The classification results for each model and the corresponding baselines are given in Table \ref{tab:results}. From Table \ref{tab:results}, overall, DGCs have higher classification performance than the corresponding baselines. The results, together with the fact that the only difference between  DGC and its baselines is the sampling layer as described in Section \ref{sec:baselines}, suggest that adding a sampling layer can improve the classification performance of a deterministic classifier.

From Table \ref{tab:results}, as for the average AUC of the 14 diseases, AlexNet encoder has the worst performance and VGGNet16 encoder have the best performance, DGC can improve the most when the encoder is AlexNet (AUC from 0.7619 to 0.7654), and improve the least when the encoder is VGG16 (AUC from 0.7875 to 0.7877). Additionally, DGC-AlexNet can consistently outperform AlexNet for all the 14 diseases. For ``Mass'', ``Pneumothorax'' and ``Consolidation'', the DGC classifiers can consistently outperform the deterministic classifiers for all the encoders. 

Horizontal comparison shows that different classification models achieve different classification performances even for the same disease. In most cases, classifiers based on VGGNet16 can outperform other types of classifiers. Out of the 14 diseases, 8 diseases achieve the best performance on VGGNet16 encoder, 5 are based on DenseNet161 and only 1 (``Hernia'') achieves the best performance on DenseNet201 encoder. From Table \ref{tab:results}, we can also see that only 2 diseases (``Cardiomegaly'' and ``Infiltration'') can achieve their best classification results on deterministic classifier (VGGNet16), all of the other diseases achieve their best performance with DGC models. 

Vertical comparison shows that the same classification model can achieve different classification performance for the 14 diseases. The most easily identified disease is ``Hernia'' (AUC=0.9113) and the least easily identified disease is ``Infiltration'' (AUC=0.6894). From Table \ref{tab:results}, the diseases that are difficult to identify include ``Infiltration'', ``Pneumonia'', ``Nodule'', ``Consolidation'' and ``Atelectasis'' (AUC$<$0.75); the easily identified diseases include ``Hernia'', ``Emphysema'' and ``Cardiomegaly'' (AUC$>$0.85).

\begin{figure*}
  \centering
  \subfloat{\includegraphics[width=0.2\textwidth,page=1]{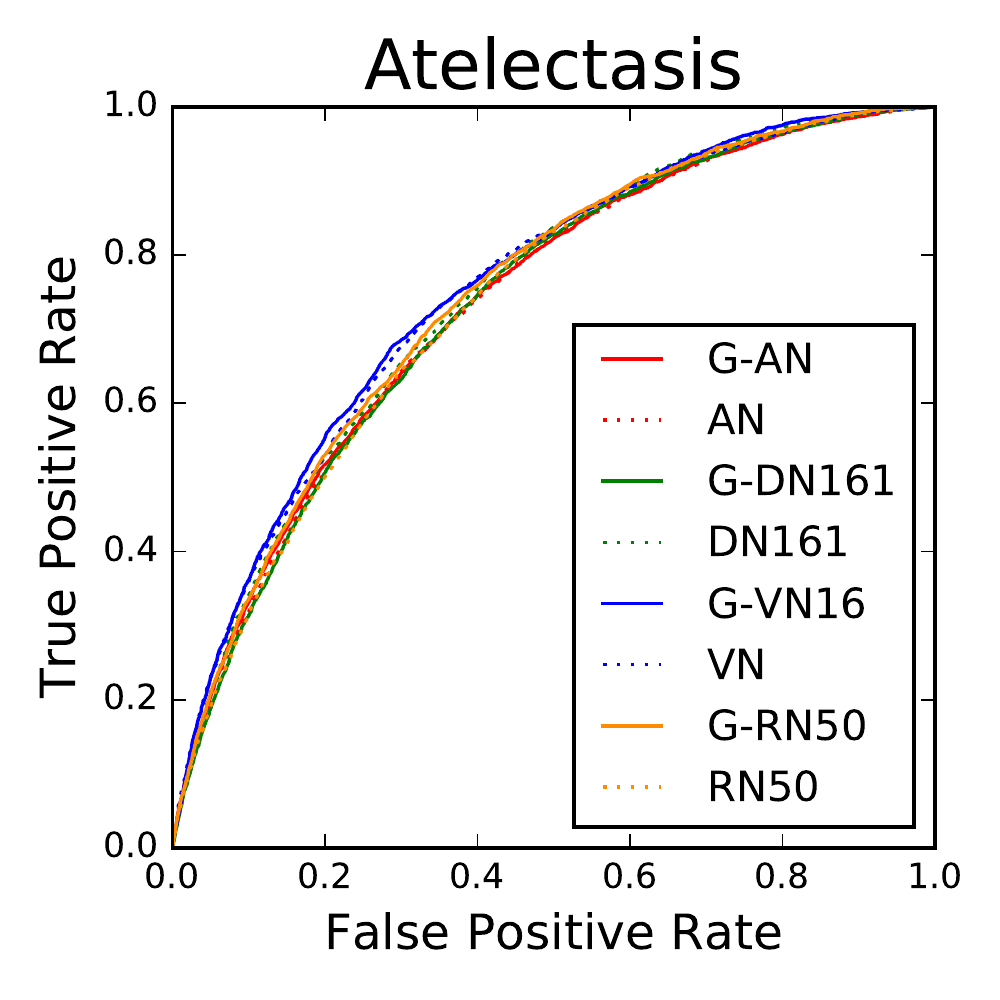}}
  \subfloat{\includegraphics[width=0.2\textwidth,page=2]{roc.pdf}}  
  \subfloat{\includegraphics[width=0.2\textwidth,page=3]{roc.pdf}}
  \subfloat{\includegraphics[width=0.2\textwidth,page=4]{roc.pdf}}
  \subfloat{\includegraphics[width=0.2\textwidth,page=5]{roc.pdf}}
  
  \subfloat{\includegraphics[width=0.2\textwidth,page=6]{roc.pdf}}  
  \subfloat{\includegraphics[width=0.2\textwidth,page=7]{roc.pdf}}
  \subfloat{\includegraphics[width=0.2\textwidth,page=8]{roc.pdf}}  
  \subfloat{\includegraphics[width=0.2\textwidth,page=9]{roc.pdf}}
  \subfloat{\includegraphics[width=0.2\textwidth,page=10]{roc.pdf}} 
  
  \subfloat{\includegraphics[width=0.2\textwidth,page=11]{roc.pdf}}
  \subfloat{\includegraphics[width=0.2\textwidth,page=12]{roc.pdf}}  
  \subfloat{\includegraphics[width=0.2\textwidth,page=13]{roc.pdf}}
  \subfloat{\includegraphics[width=0.2\textwidth,page=14]{roc.pdf}}
  \caption{ROC curves of different DGC models (i.e., DGC-AlexNet, DGC-DenseNet161, DGC-VGGNet16 and DGC-ResNet50) and their baselines over the 14 diseases.}
   \label{fig:roc}
\end{figure*}

We also plot the ROC curves of different DGC models and their baselines over the 14 diseases for one of the 5 runs, shown in Fig. \ref{fig:roc}. From Table \ref{tab:results} and Fig. \ref{fig:roc}, for most of the diseases, a DGC model can usually outperform its corresponding deterministic model. Moreover, we can see that the ROC curves of diseases ``Infiltration'', ``Pneumonia'', ``Nodule'' and ``Consolidation'' are relatively flat, which means that the classifications on these diseases are not as good as other diseases like ``Hernia'', ``Emphysema'' and ``Cardiomegaly''. This is consistent with our previous analysis.

\section{Conclusion}
In this paper, we proposed using deep generative classifiers to diagnose thoracic diseases with chest X-ray images. The deep generative classifiers contained an encoder and a classifier. The encoder encoded the input CXR image to a low-dimensional distribution, the classifier classified using features drawn from this distribution. Different from the deterministic classifiers, in the training process, generative classifiers introduce Gaussian noise and learn the variance in the training process. Through training the model with a certain amount of noise, the learned model was expected to be more robust to noise and to reduce overfitting. In this paper, we implemented the DGC architecture by adding a sampling layer between the encoder and the classifier. We used the reparameterization trick to train the DGC model through SGD. Our experimental results on ChestX-ray14 dataset demonstrated the effectiveness of the DGC models.
% * <yuan.hypnos.luo@gmail.com> 2018-09-01T21:34:58.906Z:
% 
% > Because the class label is generated from the latent distribution, it can be regarded as a generative model. 
% Again, if you are talking in terms of p(y|X), this is the classic definition of discriminative model.
% 
% ^.

The proposed DGC has similar traits with variational autoencoders (VAE). VAE is considered unsupervised, because it is trained without labels, its target is to reconstruct the original input. However, if there is a label corresponding to an image and the target is to predict the label for new images, we can solve the supervised classification problem using an adapted VAE and reconstructing the labels. This is the key idea of DGC. The difference between DGC and a deep deterministic classifier (e.g., AlexNet, VGGNet) is similar to the difference between VAE and a general autoencoder \cite{hinton1994autoencoders}.

In our architecture, we used a complex model to identify the 14 diseases, and learned the model by a sparsity-weighted cross entropy loss, while the weights for different diseases are the same, i.e., we equally regarded all the 14 diseases and learned a generic latent low-dimensional distribution for different diseases. This may somewhat influence the classification results of a certain disease. If only a certain disease is considered, one should train a specific model based on the loss corresponding to this disease. We will experiment with this setting in the future work. In addition, we did not explore the pathology localization problem using DGC, which will also be a part of our future work.

% \section*{Acknowledgment}

% The preferred spelling of the word ``acknowledgment'' in America is without 
% an ``e'' after the ``g''. Avoid the stilted expression ``one of us (R. B. 
% G.) thanks $\ldots$''. Instead, try ``R. B. G. thanks$\ldots$''. Put sponsor 
% acknowledgments in the unnumbered footnote on the first page.

\bibliographystyle{IEEEtran}
\bibliography{IEEEexample}

\begin{thebibliography}{10}
\providecommand{\url}[1]{#1}
\csname url@samestyle\endcsname
\providecommand{\newblock}{\relax}
\providecommand{\bibinfo}[2]{#2}
\providecommand{\BIBentrySTDinterwordspacing}{\spaceskip=0pt\relax}
\providecommand{\BIBentryALTinterwordstretchfactor}{4}
\providecommand{\BIBentryALTinterwordspacing}{\spaceskip=\fontdimen2\font plus
\BIBentryALTinterwordstretchfactor\fontdimen3\font minus
  \fontdimen4\font\relax}
\providecommand{\BIBforeignlanguage}[2]{{%
\expandafter\ifx\csname l@#1\endcsname\relax
\typeout{** WARNING: IEEEtran.bst: No hyphenation pattern has been}%
\typeout{** loaded for the language `#1'. Using the pattern for}%
\typeout{** the default language instead.}%
\else
\language=\csname l@#1\endcsname
\fi
#2}}
\providecommand{\BIBdecl}{\relax}
\BIBdecl

\bibitem{cdc2017pneumonia}
\BIBentryALTinterwordspacing
(2017) Pneumonia can be prevented—vaccines can help. [Online]. Available:
  \url{https://www.cdc.gov/features/pneumonia/index.html}
\BIBentrySTDinterwordspacing

\bibitem{murphy2009large}
K.~Murphy, B.~van Ginneken, A.~M. Schilham, B.~De~Hoop, H.~Gietema, and
  M.~Prokop, ``A large-scale evaluation of automatic pulmonary nodule detection
  in chest ct using local image features and k-nearest-neighbour
  classification,'' \emph{Medical image analysis}, vol.~13, no.~5, pp.
  757--770, 2009.

\bibitem{chen2011development}
S.~Chen, K.~Suzuki, and H.~MacMahon, ``Development and evaluation of a
  computer-aided diagnostic scheme for lung nodule detection in chest
  radiographs by means of two-stage nodule enhancement with support vector
  classification,'' \emph{Medical physics}, vol.~38, no.~4, pp. 1844--1858,
  2011.

\bibitem{papadopoulos2005characterization}
A.~Papadopoulos, D.~I. Fotiadis, and A.~Likas, ``Characterization of clustered
  microcalcifications in digitized mammograms using neural networks and support
  vector machines,'' \emph{Artificial Intelligence in Medicine}, vol.~34,
  no.~2, pp. 141--150, 2005.

\bibitem{domingos1996beyond}
P.~Domingos and M.~Pazzani, ``Beyond independence: Conditions for the
  optimality of the simple bayesian classi er,'' in \emph{Proc. 13th Intl.
  Conf. Machine Learning}, 1996, pp. 105--112.

\bibitem{hu2015bayesian}
B.~Hu, C.~Mao, X.~Zhang, and Y.~Dai, ``Bayesian classification with local
  probabilistic model assumption in aiding medical diagnosis,'' in
  \emph{Bioinformatics and Biomedicine (BIBM), 2015 IEEE International
  Conference on}.\hskip 1em plus 0.5em minus 0.4em\relax IEEE, 2015, pp.
  691--694.

\bibitem{wang2014non}
X.-Z. Wang, Y.-L. He, and D.~D. Wang, ``Non-naive bayesian classifiers for
  classification problems with continuous attributes,'' \emph{Cybernetics, IEEE
  Transactions on}, vol.~44, no.~1, pp. 21--39, 2014.

\bibitem{cortes1995support}
C.~Cortes and V.~Vapnik, ``Support-vector networks,'' \emph{Machine learning},
  vol.~20, no.~3, pp. 273--297, 1995.

\bibitem{fan2008liblinear}
R.-E. Fan, K.-W. Chang, C.-J. Hsieh, X.-R. Wang, and C.-J. Lin, ``Liblinear: A
  library for large linear classification,'' \emph{Journal of machine learning
  research}, vol.~9, no. Aug, pp. 1871--1874, 2008.

\bibitem{chang2011libsvm}
C.-C. Chang and C.-J. Lin, ``Libsvm: a library for support vector machines,''
  \emph{ACM transactions on intelligent systems and technology (TIST)}, vol.~2,
  no.~3, p.~27, 2011.

\bibitem{larose2006k}
D.~T. Larose and C.~D. Larose, ``k-nearest neighbor algorithm,''
  \emph{Discovering Knowledge in Data: An Introduction to Data Mining, Second
  Edition}, pp. 149--164, 2006.

\bibitem{cover1967nearest}
T.~Cover and P.~Hart, ``Nearest neighbor pattern classification,'' \emph{IEEE
  transactions on information theory}, vol.~13, no.~1, pp. 21--27, 1967.

\bibitem{mao2015nearest}
C.~Mao, B.~Hu, P.~Moore, Y.~Su, and M.~Wang, ``Nearest neighbor method based on
  local distribution for classification,'' in \emph{Advances in Knowledge
  Discovery and Data Mining}.\hskip 1em plus 0.5em minus 0.4em\relax Springer,
  2015, pp. 239--250.

\bibitem{mao2015learning}
C.~Mao, B.~Hu, M.~Wang, and P.~Moore, ``Learning from neighborhood for
  classification with local distribution characteristics,'' in \emph{Neural
  Networks (IJCNN), 2015 International Joint Conference on}.\hskip 1em plus
  0.5em minus 0.4em\relax IEEE, 2015, pp. 1--8.

\bibitem{litjens2017survey}
G.~Litjens, T.~Kooi, B.~E. Bejnordi, A.~A.~A. Setio, F.~Ciompi, M.~Ghafoorian,
  J.~A. van~der Laak, B.~Van~Ginneken, and C.~I. S{\'a}nchez, ``A survey on
  deep learning in medical image analysis,'' \emph{Medical image analysis},
  vol.~42, pp. 60--88, 2017.

\bibitem{gulshan2016development}
V.~Gulshan, L.~Peng, M.~Coram, M.~C. Stumpe, D.~Wu, A.~Narayanaswamy,
  S.~Venugopalan, K.~Widner, T.~Madams, J.~Cuadros \emph{et~al.}, ``Development
  and validation of a deep learning algorithm for detection of diabetic
  retinopathy in retinal fundus photographs,'' \emph{Jama}, vol. 316, no.~22,
  pp. 2402--2410, 2016.

\bibitem{liu2017detecting}
Y.~Liu, K.~Gadepalli, M.~Norouzi, G.~E. Dahl, T.~Kohlberger, A.~Boyko,
  S.~Venugopalan, A.~Timofeev, P.~Q. Nelson, G.~S. Corrado \emph{et~al.},
  ``Detecting cancer metastases on gigapixel pathology images,'' \emph{arXiv
  preprint arXiv:1703.02442}, 2017.

\bibitem{setio2017validation}
A.~A.~A. Setio, A.~Traverso, T.~De~Bel, M.~S. Berens, C.~van~den Bogaard,
  P.~Cerello, H.~Chen, Q.~Dou, M.~E. Fantacci, B.~Geurts \emph{et~al.},
  ``Validation, comparison, and combination of algorithms for automatic
  detection of pulmonary nodules in computed tomography images: the luna16
  challenge,'' \emph{Medical image analysis}, vol.~42, pp. 1--13, 2017.

\bibitem{zhu2017wsisa}
X.~Zhu, J.~Yao, F.~Zhu, and J.~Huang, ``Wsisa: Making survival prediction from
  whole slide histopathological images,'' in \emph{IEEE Conference on Computer
  Vision and Pattern Recognition}, 2017, pp. 7234--7242.

\bibitem{luo2017segment}
Y.~Luo, Y.~Cheng, {\"O}.~Uzuner, P.~Szolovits, and J.~Starren, ``Segment
  convolutional neural networks (seg-cnns) for classifying relations in
  clinical notes,'' \emph{Journal of the American Medical Informatics
  Association}, vol.~25, no.~1, pp. 93--98, 2017.

\bibitem{luo2017recurrent}
Y.~Luo, ``Recurrent neural networks for classifying relations in clinical
  notes,'' \emph{Journal of biomedical informatics}, vol.~72, pp. 85--95, 2017.

\bibitem{wang2017chestxray}
X.~Wang, Y.~Peng, L.~Lu, Z.~Lu, M.~Bagheri, and R.~Summers, ``Chestx-ray8:
  Hospital-scale chest x-ray database and benchmarks on weakly-supervised
  classification and localization of common thorax diseases,'' in \emph{2017
  IEEE Conference on Computer Vision and Pattern Recognition(CVPR)}, 2017, pp.
  3462--3471.

\bibitem{krizhevsky2012imagenet}
A.~Krizhevsky, I.~Sutskever, and G.~E. Hinton, ``Imagenet classification with
  deep convolutional neural networks,'' in \emph{Advances in neural information
  processing systems}, 2012, pp. 1097--1105.

\bibitem{he2016deep}
K.~He, X.~Zhang, S.~Ren, and J.~Sun, ``Deep residual learning for image
  recognition,'' in \emph{Proceedings of the IEEE conference on computer vision
  and pattern recognition}, 2016, pp. 770--778.

\bibitem{simonyan2014very}
K.~Simonyan and A.~Zisserman, ``Very deep convolutional networks for
  large-scale image recognition,'' \emph{arXiv preprint arXiv:1409.1556}, 2014.

\bibitem{huang2017densely}
G.~Huang, Z.~Liu, L.~Van Der~Maaten, and K.~Q. Weinberger, ``Densely connected
  convolutional networks.'' in \emph{CVPR}, vol.~1, no.~2, 2017, p.~3.

\bibitem{bar2015chest}
Y.~Bar, I.~Diamant, L.~Wolf, S.~Lieberman, E.~Konen, and H.~Greenspan, ``Chest
  pathology detection using deep learning with non-medical training.'' in
  \emph{ISBI}.\hskip 1em plus 0.5em minus 0.4em\relax Citeseer, 2015, pp.
  294--297.

\bibitem{donahue2014decaf}
J.~Donahue, Y.~Jia, O.~Vinyals, J.~Hoffman, N.~Zhang, E.~Tzeng, and T.~Darrell,
  ``Decaf: A deep convolutional activation feature for generic visual
  recognition,'' in \emph{International conference on machine learning}, 2014,
  pp. 647--655.

\bibitem{lakhani2017deep}
P.~Lakhani and B.~Sundaram, ``Deep learning at chest radiography: automated
  classification of pulmonary tuberculosis by using convolutional neural
  networks,'' \emph{Radiology}, vol. 284, no.~2, pp. 574--582, 2017.

\bibitem{szegedy2015going}
C.~Szegedy, W.~Liu, Y.~Jia, P.~Sermanet, S.~Reed, D.~Anguelov, D.~Erhan,
  V.~Vanhoucke, and A.~Rabinovich, ``Going deeper with convolutions,'' in
  \emph{Proceedings of the IEEE conference on computer vision and pattern
  recognition}, 2015, pp. 1--9.

\bibitem{yao2017learning}
L.~Yao, E.~Poblenz, D.~Dagunts, B.~Covington, D.~Bernard, and K.~Lyman,
  ``Learning to diagnose from scratch by exploiting dependencies among
  labels,'' \emph{arXiv preprint arXiv:1710.10501}, 2017.

\bibitem{hochreiter1997long}
S.~Hochreiter and J.~Schmidhuber, ``Long short-term memory,'' \emph{Neural
  computation}, vol.~9, no.~8, pp. 1735--1780, 1997.

\bibitem{kumar2018boosted}
P.~Kumar, M.~Grewal, and M.~M. Srivastava, ``Boosted cascaded convnets for
  multilabel classification of thoracic diseases in chest radiographs,'' in
  \emph{International Conference Image Analysis and Recognition}.\hskip 1em
  plus 0.5em minus 0.4em\relax Springer, 2018, pp. 546--552.

\bibitem{Rajpurkar2017CheXNet}
P.~Rajpurkar, J.~Irvin, K.~Zhu, B.~Yang, H.~Mehta, T.~Duan, D.~Ding, A.~Bagul,
  C.~Langlotz, and K.~Shpanskaya, ``Chexnet: Radiologist-level pneumonia
  detection on chest x-rays with deep learning,'' \emph{arXiv preprint
  arXiv:1711.05225}, 2017.

\bibitem{li2017thoracic}
Z.~Li, C.~Wang, M.~Han, Y.~Xue, W.~Wei, L.-J. Li, and F.~Li, ``Thoracic disease
  identification and localization with limited supervision,'' \emph{arXiv
  preprint arXiv:1711.06373}, 2017.

\bibitem{long2015fully}
J.~Long, E.~Shelhamer, and T.~Darrell, ``Fully convolutional networks for
  semantic segmentation,'' in \emph{Proceedings of the IEEE conference on
  computer vision and pattern recognition}, 2015, pp. 3431--3440.

\bibitem{Kingma2013auto}
D.~P. Kingma and M.~Welling, ``Auto-encoding variational bayes,'' \emph{arXiv
  preprint arXiv:1312.6114}, 2013.

\bibitem{russakovsky2015imagenet}
O.~Russakovsky, J.~Deng, H.~Su, J.~Krause, S.~Satheesh, S.~Ma, Z.~Huang,
  A.~Karpathy, A.~Khosla, M.~Bernstein \emph{et~al.}, ``Imagenet large scale
  visual recognition challenge,'' \emph{International Journal of Computer
  Vision}, vol. 115, no.~3, pp. 211--252, 2015.

\bibitem{deng2009imagenet}
J.~Deng, W.~Dong, R.~Socher, L.-J. Li, K.~Li, and L.~Fei-Fei, ``Imagenet: A
  large-scale hierarchical image database,'' in \emph{Computer Vision and
  Pattern Recognition, 2009. CVPR 2009. IEEE Conference on}.\hskip 1em plus
  0.5em minus 0.4em\relax Ieee, 2009, pp. 248--255.

\bibitem{ioffe2015batch}
S.~Ioffe and C.~Szegedy, ``Batch normalization: accelerating deep network
  training by reducing internal covariate shift,'' in \emph{Proceedings of the
  32nd International Conference on International Conference on Machine
  Learning-Volume 37}.\hskip 1em plus 0.5em minus 0.4em\relax JMLR. org, 2015,
  pp. 448--456.

\bibitem{boureau2010theoretical}
Y.-L. Boureau, J.~Ponce, and Y.~LeCun, ``A theoretical analysis of feature
  pooling in visual recognition,'' in \emph{Proceedings of the 27th
  international conference on machine learning (ICML-10)}, 2010, pp. 111--118.

\bibitem{sutskever2013importance}
I.~Sutskever, J.~Martens, G.~Dahl, and G.~Hinton, ``On the importance of
  initialization and momentum in deep learning,'' in \emph{International
  conference on machine learning}, 2013, pp. 1139--1147.

\bibitem{kingma2014adam}
D.~P. Kingma and J.~L. Ba, ``Adam: Amethod for stochastic optimization,'' in
  \emph{Proc. 3rd Int. Conf. Learn. Representations}, 2014.

\bibitem{he2015delving}
K.~He, X.~Zhang, S.~Ren, and J.~Sun, ``Delving deep into rectifiers: Surpassing
  human-level performance on imagenet classification,'' in \emph{Proceedings of
  the IEEE international conference on computer vision}, 2015, pp. 1026--1034.

\bibitem{hinton1994autoencoders}
G.~E. Hinton and R.~S. Zemel, ``Autoencoders, minimum description length and
  helmholtz free energy,'' in \emph{Advances in neural information processing
  systems}, 1994, pp. 3--10.

\end{thebibliography}

\end{document}